# Deep Reinforcement Learning for Robotic Manipulation with Asynchronous Off-Policy Updates

Shixiang Gu[*,1,2,3] and Ethan Holly[*,1] and Timothy Lillicrap[4] and Sergey Levine[1,5]

*Abstract*— Reinforcement learning holds the promise of enabling autonomous robots to learn large repertoires of behavioral skills with minimal human intervention. However, robotic applications of reinforcement learning often compromise the autonomy of the learning process in favor of achieving training times that are practical for real physical systems. This typically involves introducing hand-engineered policy representations and human-supplied demonstrations. Deep reinforcement learning alleviates this limitation by training general-purpose neural network policies, but applications of direct deep reinforcement learning algorithms have so far been restricted to simulated settings and relatively simple tasks, due to their apparent high sample complexity. In this paper, we demonstrate that a recent deep reinforcement learning algorithm based on off-policy training of deep Q-functions can scale to complex 3D manipulation tasks and can learn deep neural network policies efficiently enough to train on real physical robots. We demonstrate that the training times can be further reduced by parallelizing the algorithm across multiple robots which pool their policy updates asynchronously. Our experimental evaluation shows that our method can learn a variety of 3D manipulation skills in simulation and a complex door opening skill on real robots without any prior demonstrations or manually designed representations.

## I. INTRODUCTION

Reinforcement learning methods have been applied to range of robotic control tasks, from locomotion [1], [2] to manipulation [3], [4], [5], [6] and autonomous vehicle control [7]. However, practical real-world applications of reinforcement learning have typically required significant additional engineering beyond the learning algorithm itself: an appropriate representation for the policy or value function must be chosen so as to achieve training times that are practical for physical hardware [8], and example demonstrations must often be provided to initialize the policy and mitigate safety concerns during training [9]. In this work, we show that a recently proposed deep reinforcement learning algorithms based on off-policy training of deep Q-functions [10], [11] can be extended to learn complex manipulation policies from scratch, without user-provided demonstrations, and using only general-purpose neural network representations that do not require task-specific domain knowledge.

One of the central challenges with applying direct deep reinforcement learning algorithms to real-world robotic platforms has been their apparent high sample-complexity. We demonstrate that, contrary to commonly held assumptions, recently developed off-policy deep Q-function based algorithms such as the Deep Deterministic Policy Gradient algorithm (DDPG) [10] and Normalized Advantage Function algorithm (NAF) [11] can achieve training times that are suitable for real robotic systems. We also demonstrate that we can further reduce training times by parallelizing the algorithm across multiple robotic platforms. To that end, we present a novel asynchronous variant of NAF, evaluate the speedup obtained with varying numbers of learners in simulation, and demonstrate real-world results with parallelism across multiple robots. An illustration of these robots learning a door opening task is shown in Figure 1.

[*]equal contribution, [1]Google Brain, [2]University of Cambridge, [3]MPI Tübingen, [4]Google DeepMind, [5]UC Berkeley

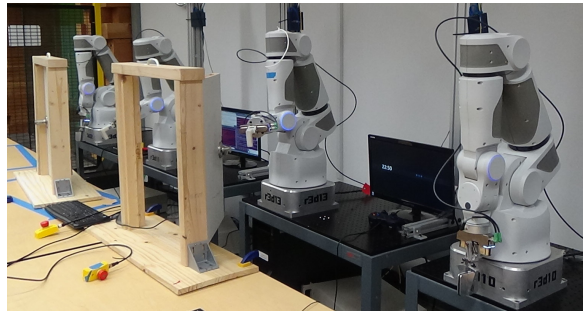

Fig. 1: Two robots learning a door opening task. We present a method that allows multiple robots to cooperatively learn a single policy with deep reinforcement learning.

The main contribution of this paper is a demonstration of asynchronous deep reinforcement learning using our parallel NAF algorithm across a cluster of robots. Our technical contribution consists of the asynchronous variant of the NAF algorithm, as well as practical extensions of the method to enable sample-efficient training on real robotic platforms. We also introduce a simple and effective safety mechanism for constraining exploration at training time, and present simulated experiments that evaluate the speedup obtained from parallelizing across a variable number of learners. Our experiments also evaluate the benefits of deep neural network representations for several complex manipulation tasks, including door opening and pick-and-place, by comparing to more standard linear representations. Our real world experiments show that our approach can be used to learn a door opening skill from scratch using only general-purpose neural network representations and without any human demonstrations. To the best of our knowledge, this is the first demonstration of autonomous door opening that does not use human-provided examples for initialization.

## II. RELATED WORK

Applications of reinforcement learning (RL) in robotics have included locomotion [1], [2], manipulation [3], [4],

[5], [6], and autonomous vehicle control [7]. Many of the RL methods demonstrated on physical robotic systems have used relatively low-dimensional policy representations, typically with under one hundred parameters, due to the difficulty of efficiently optimizing high-dimensional policy parameter vectors [12]. Although there has been considerable research on reinforcement learning with general-purpose neural networks for some time [13], [14], [15], [16], [17], such methods have only recently been developed to the point where they could be applied to continuous control of high-dimensional systems, such as 7 degree-of-freedom (DoF) arms, and with large and deep neural networks [18], [19], [10], [11]. This has made it possible to learn complex skills with minimal manual engineering, though it has remained unclear whether such approaches could be adapted to real systems given their sample complexity.

In robotic learning scenarios, prior work has explored both model-based and model-free learning algorithms. Model-based algorithms have explored a variety of dynamics estimation schemes, including Gaussian processes [20], mixture models [21], and local linear system estimation [22], with a more detailed overview in a recent survey [23]. Deep neural network policies have been combined with model-based learning in the context of guided policy search algorithms [19], which use a model-based teacher to train a deep network policies. Such methods have been successful on a range of real-world tasks, but rely on the ability of the model-based teacher to discover good trajectories for the goal task. As shown in recent work, this can be difficult in domains with severe discontinuities in the dynamics and reward function [24].

In this work, we focus on model-free reinforcement learning, which includes policy search methods [25], [3], [23] and function approximation methods [26], [27], [14]. Both approaches have recently been combined with deep neural networks for learning complex tasks [28], [29], [18], [10], [11], [30]. Direct policy gradient methods offer the benefit of unbiased gradient estimates, but tend to require more experience, since on-policy estimators preclude reuse of past data [18]. On the other hand, methods based on Q-function estimation often allow us to leverage past off-policy data. We therefore build on Q-function based methods, extending Normalized Advantage Estimation (NAF) [11], which extends Q-learning to continuous action spaces. This method is closely related to Deep Deterministic Policy Gradient (DDPG) [10] as well as NFQCA [31], with principle differences being that NFQCA employs a batch episodic update and typically resets network parameters between episodes.

Accelerating robotic learning by pooling experience from multiple robots has long been recognized as a promising direction in the domain of cloud robotics, where it is typically referred to as collective robotic learning [32], [33], [34], [35]. In deep reinforcement learning, parallelized learning has also been proposed to speed up simulated experiments [30]. The goals of this prior work are fundamentally different from ours: while prior asynchronous deep reinforcement learning work seeks to reduce overall training time, under the assumption that simulation time is inexpensive and the training is dominated by neural network computations, our work instead seeks to minimize the training time when training on real physical robots, where experience is expensive and computing neural network backward passes is comparatively cheap. In this case, we retain the use of a replay buffer, and focus on asynchronous execution and neural network training. Our results demonstrate that we achieve significant speedup in overall training time from simultaneously collecting experience across multiple robotic platforms. This result agrees with a concurrent result on parallelized guided policy search [36], and in fact we tackle a similar door opening task. In contrast to this concurrent work, we perform the task from scratch, without initializing the algorithm from user demonstrations. While the concurrent work by Yahya et al. provides a more extensive exploration of generalization to physical variation, we focus on analyzing learning speed of model-free deep reinforcement learning methods.

III. BACKGROUND

In this section, we will formulate the robotic reinforcement learning problem, introduce essential notation, and describe the existing algorithmic foundations on which we build the methods for this work. The goal in reinforcement learning is to control an agent attempting to maximize a reward function which, in the context of a robotic skill, denotes a user-provided definition of what the robot should try to accomplish. At state $x_t$ in time $t$, the agent chooses and executes action $u_t$ according to its policy $\pi(u_t|x_t)$, transitions to a new state $x_t$ according to the dynamics $p(x_t|x_t, u_t)$ and receives a reward $r(x_t, u_t)$. Here, we consider infinite-horizon discounted return problems, where the objective is the $\gamma-$discounted future return from time $t$ to $\infty$, given by $R_t = \sum_{i=t}^{\infty} \gamma^{(i-t)} r(x_i, u_i)$. The goal is to find the optimal policy $\pi^*$ which maximizes the expected sum of returns from the initial state distribution, given by $R = \mathbb{E}_\pi[R_1]$.

Among reinforcement learning methods, off-policy methods such as Q-learning offer significant data efficiency compared to on-policy variants, which is crucial for robotics applications. Q-learning trains a greedy deterministic policy $\pi(u_t|x_t) = \delta(u_t = \mu(x_t))$ by iterating between learning the Q-function, $Q^{\pi_n}(x_t, u_t) = \mathbb{E}_{r_{i\geq t}, x_{i>t} \sim E, u_{i>t} \sim \pi_n}[R_t|x_t, u_t]$, of a policy and updating the policy by greedily maximizing the Q-function, $\mu_{n+1}(x_t) = \arg\max_u Q^{\pi_n}(x_t, u_t)$. Let $\theta^Q$ parametrize the action-value function, $\beta$ be an arbitrary exploration policy, and $\rho^\beta$ be the state visitation distribution induced by $\beta$, the learning objective is to minimize the Bellman error, where we fix the target $y_t$:

$$L(\theta^Q) = \mathbb{E}_{x_t \sim \rho^\beta, u_t \sim \beta, x_{t+1}, r_t \sim E}[(Q(x_t, u_t|\theta^Q) - y_t)^2]$$
$$y_t = r(x_t, u_t) + \gamma Q(x_{t+1}, \mu(x_{t+1}))$$

For continuous action problems, the policy update step is intractable for a Q-function parametrized by a deep neural network. Thus, we will investigate Deep Deterministic Policy Gradient (DDPG) [10] and Normalized Advantage Functions (NAF) [11]. DDPG circumvents the problem by adopting an actor-critic method, while NAF restricts the class of

Q-function to the expression below to enable closed-form updates, as in the discrete action case. During exploration, a temporally-correlated noise is added to the policy network output. For more details and comparisons on DDPG and NAF, please refer to [10], [11] as well as experimental results in Section V-B.

$$Q(\boldsymbol{x},\boldsymbol{u}|\theta^Q) = A(\boldsymbol{x},\boldsymbol{u}|\theta^A) + V(\boldsymbol{x}|\theta^V)$$
$$A(\boldsymbol{x},\boldsymbol{u}|\theta^A) = -\frac{1}{2}(\boldsymbol{u}-\boldsymbol{\mu}(\boldsymbol{x}|\theta^\mu))^T \boldsymbol{P}(\boldsymbol{x}|\theta^P)(\boldsymbol{u}-\boldsymbol{\mu}(\boldsymbol{x}|\theta^\mu))$$

We evaluate both DDPG and NAF in our simulated experiments, where they yield comparable performance, with NAF producing slightly better results overall for the tasks examined here. On real physical systems, we focus on variants of the NAF method, which is simpler, requires only a single optimization objective, and has fewer hyper-parameters.

This RL formulation can be applied on robotic systems to learn a variety of skills defined by reward functions. However, the learning process is typically time consuming, and requires a number of practical considerations. In the next section, we will present our main technical contribution, which consists of a parallelized variant of NAF, and also discuss a variety of technical contributions necessary to apply NAF to real-world robotic skill learning.

## IV. ASYNCHRONOUS TRAINING OF NORMALIZED ADVANTAGE FUNCTIONS

In this section, we present our primary contribution: an extension of NAF that makes it practical for use with real-world robotic platforms. To that end, we describe how online training of the Q-function estimator can be performed asynchronously, with a learner thread that trains the network and one or more worker threads that collect data by executing the current policy on one or more robots. Besides making NAF suitable for real time applications, this approach also makes it straightforward to collect experience from multiple robots in parallel. This is crucial in real-world robot learning, since the learning time is often constrained by the data collection rate in real time, rather than network training speed. When data collection is the limiting factor, then 2-3 times quicker data collection may translate directly to 2-3 times faster skill acquisition on a real robot. We also describe practical considerations, such as safety constraints, which are necessary in order to allow the exploration required to train complex policies from scratch on real systems. To the best of our knowledge, this is the first direct deep RL method that has been demonstrated on a real robotics platform with many DoFs and contact dynamics, and without demonstrations or simulated pretraining [18], [10], [11]. As we will show in our experimental evaluation, this approach can be used to learn complex tasks such as door opening from scratch, which previously required additional details such as human demonstrations to succeed [6].

### A. Asynchronous Learning

In asynchronous NAF, the learner thread is separated from the experience collecting worker threads. The asynchronous

**Algorithm 1** Asynchronous NAF - $N$ collector threads and 1 trainer thread

// trainer thread
Randomly initialize normalized Q network $Q(\boldsymbol{x},\boldsymbol{u}|\theta^Q)$, where $\theta^Q = \{\theta^\mu, \theta^P, \theta^V\}$ as in Eq. 1
Initialize target network $Q'$ with weight $\theta^{Q'} \leftarrow \theta^Q$
Initialize shared replay buffer $R \leftarrow \emptyset$
**for** iteration=1,$I$ **do**
　　Sample a random minibatch of $m$ transitions from $R$
　　Set $y_i = \begin{cases} r_i + \gamma V'(\boldsymbol{x}'_i|\theta^{Q'}) & \text{if} \quad t_i < T \\ r_i & \text{if} \quad t_i = T \end{cases}$
　　Update the weight $\theta^Q$ by minimizing the loss: $L = \frac{1}{m}\sum_i(y_i - Q(\boldsymbol{x}_i,\boldsymbol{u}_i|\theta^Q))^2$
　　Update the target network: $\theta^{Q'} \leftarrow \tau\theta^Q + (1-\tau)\theta^{Q'}$
**end for**
// collector thread $n$, $n = 1...N$
Randomly initialize policy network $\boldsymbol{\mu}(\boldsymbol{x}|\theta^\mu_n)$
**for** episode=1,$M$ **do**
　　Sync policy network weight $\theta^\mu_n \leftarrow \theta^\mu$
　　Initialize a random process $\mathcal{N}$ for action exploration
　　Receive initial observation state $\boldsymbol{x}_1 \sim p(\boldsymbol{x}_1)$
　　**for** t=1,$T$ **do**
　　　　Select action $\boldsymbol{u}_t = \boldsymbol{\mu}(\boldsymbol{x}_t|\theta^\mu_n) + \mathcal{N}_t$
　　　　Execute $\boldsymbol{u}_t$ and observe $r_t$ and $\boldsymbol{x}_{t+1}$
　　　　Send transition $(\boldsymbol{x}_t,\boldsymbol{u}_t,r_t,\boldsymbol{x}_{t+1},t)$ to $R$
　　**end for**
**end for**

learning algorithm is summarized in Algorithm 1. The learner thread uses the replay buffer to perform asynchronous updates to the deep neural network Q-function approximator. This thread runs on a central server, and dispatches updated policy parameters to each of the worker threads. The experience collecting worker threads run on the individual robots, and send the observation, action, and reward for each time step to the central server to append to the replay buffer. This decoupling between the training and the collecting threads allows the controllers on each of the robots to run in real time, without experiencing delays due to the computational cost of backpropagation through the network. Furthermore, it makes it straightforward to parallelize experience collection across multiple robots simply by adding additional worker threads. We only use one thread for training the network; however, the gradient computation can also be distributed in same way as [30] within our framework. While the trainer thread keeps training from the centralized replay buffer, the collector threads sync their policy parameters with the trainer thread at the beginning of each episode, execute commands on the robots, and push experience into the buffer.

### B. Safety Constraints

Ensuring safe exploration poses a significant challenge for real-world training with reinforcement learning. Q-learning requires a significant amount of noisy exploration for gathering the experience necessary for action-value function approximation. For all experiments, we set a maximum commanded velocity allowed per joint, as well as strict position limits for each joint. In addition to joint position limits, we used a bounding sphere for the end-effector position. If the commanded joint velocities would send the end-effector outside of the sphere, we used the forward kinematics to project the commanded velocity onto the surface of the sphere, plus some correction velocity to force toward the center. For experiments with no contacts, these safety constraints were sufficient to prevent unsafe exploration; for experiments with contacts, additional heuristics were required for safety.

*C. Network Architectures*

To minimize manual engineering, we use a simple and readily available state representation consisting of joint angles and end-effector positions, as well as their time derivatives. In addition, we append a target position to the state, which depends on the task: for the reaching task, this is the goal position for the end-effector; for the door opening, this is the handle position when the door is closed and the quaternion measurement of the sensor attached to the door frame. Since the state representation is compact, we use standard feed-forward networks to parametrize the action-value functions and policies. We use two-hidden-layer network with size 100 units each to parametrize each of $\boldsymbol{\mu}(x)$, $\boldsymbol{L}(x)$ (Cholesky decomposition of $\boldsymbol{P}(x)$), and $V(x)$ in NAF and $\boldsymbol{\mu}(x)$ and $Q(x,\boldsymbol{u})$ in DDPG. For $Q(x,\boldsymbol{u})$ in DDPG, the action vector $\boldsymbol{u}$ added as another input to second hidden layer followed by a linear projection. ReLU is used as hidden activations and hyperbolic tangent (Tanh) is used for the final layer activation function in the policy networks $\boldsymbol{\mu}(x)$ to bound the action scale.

To illustrate the importance of deep neural networks for representing policies or action-value functions, we study these neural network models against another simpler parametrization. Specifically we study a variant of NAF (Linear-NAF) as below, where $\boldsymbol{\mu}(x) = f(\boldsymbol{k}+\boldsymbol{K}x)$, $\boldsymbol{P},\boldsymbol{k},\boldsymbol{K},\boldsymbol{B},\boldsymbol{b},c$ are learnable matrices, vectors, or scalars of appropriate dimension, and $f$ is Tanh to enforce bounded actions.

$$Q(\boldsymbol{x},\boldsymbol{u}) = \frac{1}{2}(\boldsymbol{u}-\boldsymbol{\mu}(\boldsymbol{x}))^T \boldsymbol{P}(\boldsymbol{u}-\boldsymbol{\mu}(\boldsymbol{x})) + \boldsymbol{x}^T\boldsymbol{B}\boldsymbol{x} + \boldsymbol{x}^T\boldsymbol{b} + c$$

If $f$ is identity, then the expression corresponds to a globally quadratic Q-function and a linear feedback policy, though due to the Tanh non-linearity, the Q-function is not linear with respect to state-action features.

## V. SIMULATED EXPERIMENTS

We first performed a detailed investigation of the learning algorithms using simulated tasks modeled using the MuJoCo physics simulator [37]. Simulated environments enable fast comparisons of design choices, including update frequencies, parallelism, network architectures, and other hyperparameters. We modeled a 7-DoF lightweight arm that was also used in our physical robot experiments, as well as a 6-DoF Kinova JACO arm with 3 additional degrees of freedom in the fingers, for a total of 9 degrees of freedom. Both arms were controlled at the level of joint velocities, except the three JACO finger joints which are controlled with torque actuators. The 7-DoF arm is controlled at 20Hz to match the real-world robot experiments, and the JACO arm is controlled at 100Hz. Gravity is turned off for the 7-DoF arm, which is a valid assumption given that the actual robot uses built-in gravity compensation. Gravity is enabled for the JACO arm. Different arm geometries, control frequencies, and gravity settings illustrate the learning algorithm's robustness to different learning environments.

*A. Simulation Tasks*

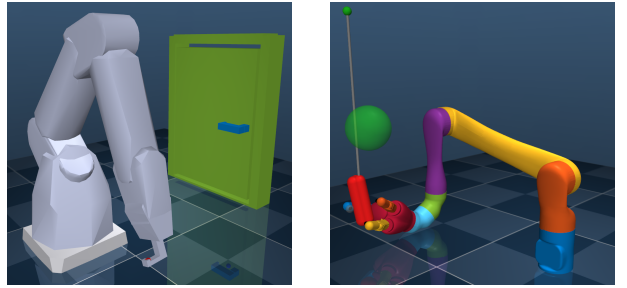

Fig. 2: The 7-DoF arm and JACO arm in simulation.

Tasks include random-target reaching, door pushing, door pulling, and pick & place in a 3D environment, as detailed below. The 7-DoF arm is set up for the random target reaching and door tasks, while the JACO arm is used for the pick & place task (see Figure 2). Details of each task are below, where $d$ is Huber loss and $c_i$'s are non-negative constants. Discount factor of $\gamma = 0.98$ is chosen and the Adam optimizer [38] with base learning rate of either 0.0001 or 0.001 is used for all the experiments. Importantly, almost no hyperparameter search was required to ensure that the employed algorithms were successful across robot and task.

*1) Reaching (7-DoF arm):* The 7-DoF arm tries to reach a random target in space from a fixed initial configuration. A random target is generated per episode by sampling points uniformly from a cube of size 0.2m centered around a point. State features include the 7 joint angles and their time derivatives, the end-effector position and the target position, totalling 20 dimensions. Each episode duration is 150 time steps (7.5 seconds). Success rate is computed from 5 random test episodes where an episode is successful if the arm can reach within 5 cm of the target. Given the end-effector position $\boldsymbol{e}$ and the target position $\boldsymbol{y}$, the reward function is below,

$$r(\boldsymbol{x},\boldsymbol{u}) = -c_1 d(\boldsymbol{y}, \boldsymbol{e}(\boldsymbol{x})) - c_2 \boldsymbol{u}^T \boldsymbol{u}$$

*2) Door Pushing and Pulling (7-DoF arm):* The 7-DoF arm tries to open the door by pushing or pulling the handle (see Figure 2). For each episode, the door position is sampled randomly within a rectangle of 0.2m by 0.1m. The handle can be turned downward for up to 90 degrees, while the door can be opened up to 90 degrees in both directions. The door has a spring such that it closes gradually when no force is applied. The door has a latch such that it could only open the door only when the handle is turned past approximately 60 degrees. To make the setting similar to the real robot experiment where the quaternion readings from the

VectorNav IMU are used for door angle measurements, the quaternion of the door handle is used to compute the loss. The reward function is composed of two parts: the closeness of the end-effector to the handle, and the measure of how much the door is opened in the right direction. The first part depends on the distance between end-effector position $e$ and the handle position $h$ in its neutral state. The second part depends on the distance between the quaternion of the handle $q$ and its value when the handle is turned and door is opened $q_o$. We also added the distance when the door is at neutral position as offset $d_i = d(q_o, q_i)$ such that, when the door is opened the correct way, it receives positive reward. State features include the 7 joint angles and their time derivatives, the end-effector position, the resting handle position, the door frame position, the door angle, and the handle angle, totally 25 dimensions. Each episode duration is 300 time steps (15 seconds). Success rate is computed from 20 random test episodes where an episode is successful if the arm can open the door in the correct direction by a minimum of 10 degrees.

$$r(x,u) = -c_1 d(h, e(x)) + c_2 (-d(q_o, q(x)) + d_i) - c_3 u^T u$$

*3) pick & place (JACO):* The JACO arm tries to pick up a stick suspending in the air by a string and place it near the target upward in the space (see Figure 2). The hand begins near to, but not in contact with the stick, so the grasp must be learned. The task is similar to a task previously explored with on-policy methods [30], except that here the task requires moving the stick to multiple targets. For each episode a new target is sampled from a square of size 0.24 m at a fixed height, while the initial stick position and the arm configuration are fixed. Given the grip site position $g$ (where the three fingers meet when closed), the three finger tip positions $f_1, f_2, f_3$, the stick position $s$ and the target position $y$, the reward function is below. State features include the position and rotation matrices of all geometries in the environment, the target position and the vector from the stick to the target, totally 180 dimensions. The large observation dimensionality creates an interesting comparison with the above two tasks. Each episode duration is 300 time steps (3 seconds). Success rate is computed from 20 random test episodes where an episode is judged successful if the arm can bring the stick within 5 cm of the target.

$$r(x,u) = -c_1 d(s(x), g(x)) - c_2 \sum_{i=1}^{3} d(s(x), f_i(x)) \\ - c_3 d(y, s(x)) - c_4 u^T u$$

### B. Neural Network Policy Representations

Neural networks are powerful function approximators, but they have significantly more parameters than the simpler linear models that are often used in robotic learning [23], [8]. In this section, we compare empirical performance of DDPG, NAF, and Linear-NAF as described in Section IV-C. In particular, we want to verify if deep representations for policy and value functions are necessary for solving complex tasks from scratch, and evaluate how they compare with linear models in terms of convergence rate. For the 7-DoF arm tasks, DDPG and NAF models have significantly more parameters than Linear-NAF, while the pick & place task has a high-dimensional observation, and thus the parameter sizes are more comparable. Of course, many other linear representations are possible, including DMPs [39], splines [3], and task-specific representations [40]. This comparison only serves to illustrate that our tasks are complex enough that simple, fully generic linear representations are not by themselves sufficient for success. For the experiments in this section, batch normalization [41] is applied. These experiments were conducted synchronously, where 1 parameter update is applied per 1 time step in simulation.

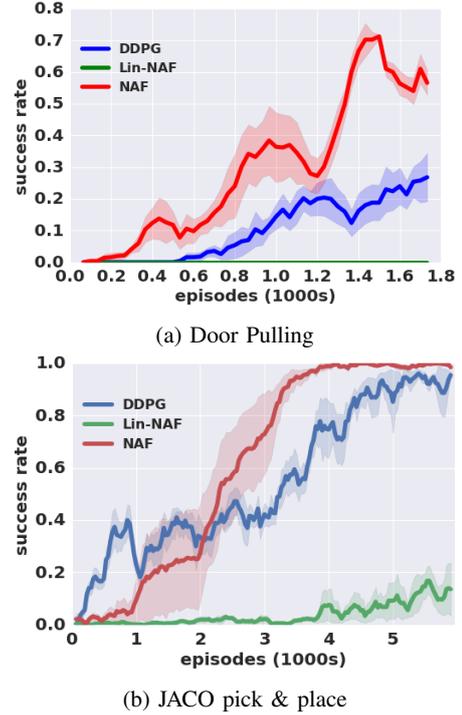

(a) Door Pulling

(b) JACO pick & place

Fig. 3: The figure shows the learning curves for two tasks, comparing DDPG, Linear-NAF, and NAF. Note that the linear model struggles to learn the tasks, indicating the importance of expressive nonlinear policy representations.

Figure 3 shows the experimental results on the 7-DoF door pulling and JACO pick & place tasks and Table 4 summarizes the overall results. For reaching and pick & place, Linear-NAF learns good policies competitive with those of NAF and DDPG, but converges significantly slower than both NAF and DDPG. This is contrary to common belief that neural networks take significantly more data and update steps to converge to good solutions. One possible explanation is that in RL the data collection and the model learning are coupled, and if the model is more expressive, it can explore a greater variety of complex policies efficiently and thus collect diverse and good data quickly. This is not a problem for well-pre-trained policy learning but could be an important issue when learning from scratch. In the case of door tasks, the linear model completely fails to learn perfect policies. More thorough investigations into how expressivity of the policy

|             | Max. success rate (%) |         |         | Episodes to 100% success (1000s) |        |         |
|-------------|-----------------------|---------|---------|----------------------------------|--------|---------|
|             | DDPG                  | Lin-NAF | NAF     | DDPG                             | Lin-NAF| NAF     |
| Reach       | 100±0                 | 100±0   | 100±0   | 3.2±0.7                          | 8±3    | 3.6±1.0 |
| Door Pull   | 100±0                 | 5±6     | 100±0   | 10±8                             | N/A    | 6±3     |
| Door Push   | 100±0                 | 40±10   | 100±0   | 3.1±1.0                          | N/A    | 4.2±1.0 |
| Pick & Place| 100±0                 | 100±0   | 100±0   | 4.4±0.6                          | 12±3   | 2.9±0.9 |

Fig. 4: The table summarizes the performances of DDPG, Linear-NAF, and NAF across four tasks. Note that the linear model learns the perfect reaching and pick & place policies given enough time, but fails to learn either of the door tasks.

interact with reinforcement learning is a promising direction for future work.

Additionally, the experimental results on the door tasks show that Linear-NAF does not succeed in learning such tasks. The difference from above tasks likely comes from the complexity of policies. For reaching and pick & place, the tasks mainly requires learning single-motion policies, e.g. close fingers to grasp the stick and move it to the target. For the door tasks, the robot is required to learn how to hook onto the door handle in different locations, turn it, and push or pull. See the supplementary video at https://sites.google.com/site/deeproboticmanipulation/ for learned resulting behaviors for each tasks.

*C. Asynchronous Training*

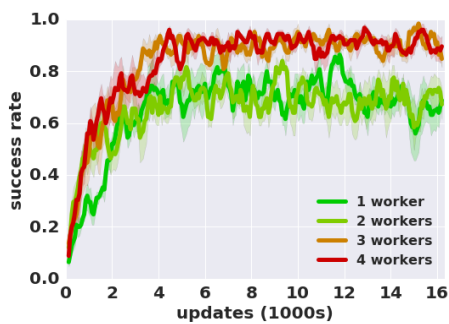

(a) Reaching

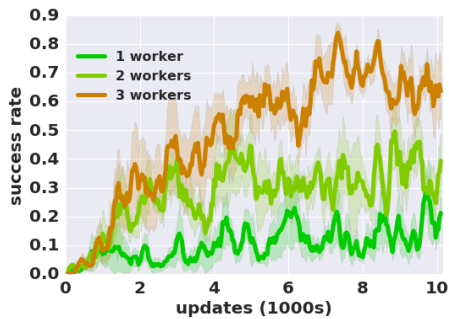

(b) Door Pushing

Fig. 5: Asynchronous training of NAF in simulation. Note that both learning speed and final policy success rates depending significantly on the number of workers.

In asynchronous training, the training thread continuously trains the network at a fixed frequency determined by network size and computational hardware, while each collector thread runs at a specified control frequency. The main question to answer is: given these constraints, how much speedup can we gain from increasing the number of workers, i.e. the data collection speed? To analyze this in a realistic but controlled setting, we first set up the following experiment in simulation. We locked each collector thread to run at $S$ times the speed of the training thread. Then, we varied the number of collector threads $N$. Thus, the overall data collection speed is approximately $S \times N$ times that of the trainer thread. For our experiments, we varied $N$ and fixed $S = 1/5$ since our training thread runs at approximately 100 updates per second on CPU, while the collector thread in real robot will be locked to 20Hz. Layer normalization is applied [42].

Figure 5 shows the results on reaching and door pushing. The x-axis shows the number of parameter updates, which is proportional to the amount of wall-clock time required for training, since the amount of data per step increases with the number of workers. The results demonstrate three points: (1) under some circumstances, increasing data collection makes the learning converge significantly faster with respect to the number of gradient steps, (2) final policy performances depend a lot on the ratio between collecting and training speeds, and (3) there is a limit where collecting more data does not help speed up learning. However, we hypothesize that accelerating the speed of neural network training, which in these cases was pegged to one update per time step, could allow the model to ingest more data and benefit more from greater parallelism. This is particularly relevant as parallel computational hardware, such as GPUs, are improved and deployed more widely. Videos of the learned policies are available in supplementary materials and online: https://sites.google.com/site/deeproboticmanipulation/

VI. REAL-WORLD EXPERIMENTS

The real-world experiments are conducted with the 7-DoF arm shown in Figure 6. The tasks are the same as the simulation tasks in Section V-A with some minor changes. For reaching, the same state representation and reward functions are used. The randomized target position is sampled from a cube of 0.4 m, providing more diverse and extreme targets for reaching. We noticed that these more aggressive targets, combined with stricter safety measures (slower movements and tight joint limits), reduced the performance compared to the simulation, and thus we relax the definition of a successful episode for reporting, marking episodes within 10 cm as successful. For the door task, the robot was required to reach for and pull the door open by hooking the handle with the end-effector. Due to the geometry of the workspace, we could not test the door pushing task on the real hardware. The orientation of the door was measured by a VectorNav IMU attached to the back of the door. Unlike in the simulation, we cannot automatically reposition the door for every episode, so the pose of the door was kept fixed. State features for the door task include the joint angles and their time derivatives, the end effector position and the quaternion reading from the IMU, totalling 21 dimensions.

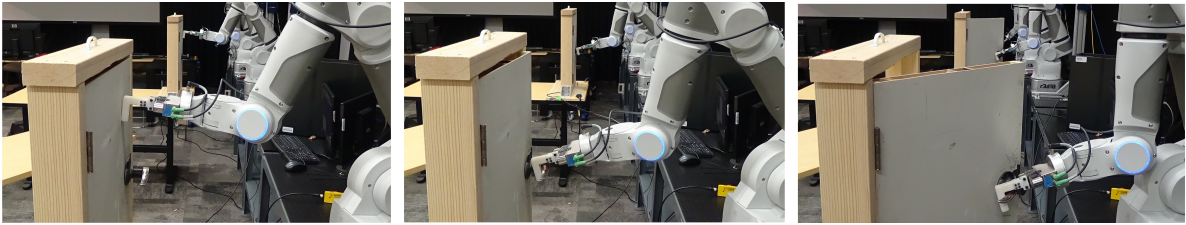

Fig. 6: Two robots learning to open doors using asynchronous NAF. The final policy learned with two workers could achieve a 100% success rate on the task across 20 consecutive trials.

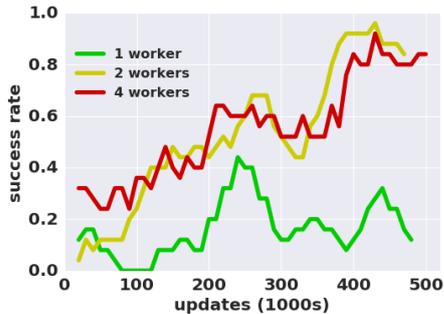

Fig. 7: The 7-DoF arm random target reaching with asynchronous NAF on real robots. Note that 1 worker suffers in both learning speed and final policy performance.

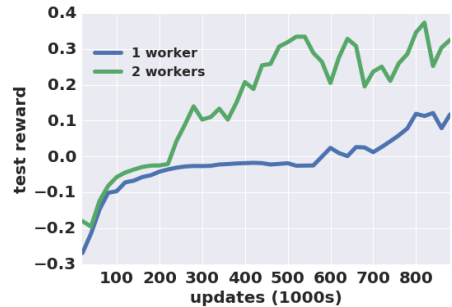

Fig. 8: Learning curves for real-world door opening. Learning with two workers significantly outperforms the single worker, and achieves a 100% success rate in under 500,000 update steps, corresponding to about 2.5 hours of real time.

### A. Random Target Reaching

The simulation results in Section 5 provide approximate performance gains that can be expected from parallelism. However, the simulation setting does not consider several issues that could arise in real-world experiments: delays due to slow resetting procedures, non-constant execution speeds of the training thread, and subtle physics discrepancies among robots. Thus, it is important to demonstrate the benefits from parallel training with real robots.

We set up the same reaching experiment in the real world across up to four robots. Robots execute policies at 20 Hz, while the training thread simply updates the network continuously at approximately 100 Hz. The same network architecture and hyper-parameters from the simulation experiment are used.

Figure 7 confirms that 2 or 4 workers significantly improves learning speed over 1 worker, though the gains on this simple task are not substantial past 2 workers. Importantly, when the training thread is not synchronized with the data collection thread and the data collection is too slow, it may not just slow down learning but also hurt the final policy performance, as observed in the 1-worker case. Further discrepancies from the simulation may also be explained by physical discrepancies among different robots. The learned policies are presented in the supplementary video.

### B. Door Opening

The previous section describes a real-world evaluation of asynchronous NAF and demonstrates that learning can be accelerated by using multiple workers. In this section, we describe a more complex door opening task. Door opening presents a practical application of robotic learning that involves complex and discontinuous contact dynamics. Previous work has demonstrated learning of door opening policies using example demonstration provided by a human expert [6]. In this work, we demonstrate that we can learn policies for pulling open a door from scratch using asynchronous NAF. The entire task required approximately 2.5 hours to learn with two workers learning simultaneously, and the final policy achieves 100% success rate evaluated across 20 consecutive trials. An illustration of this task is shown in Figure 6, and the supplementary video shows different stages in the learning process, as well as the final learned policy.

Figure 8 illustrates the difference in the learning process between one and two workers, where the horizontal axis shows the number of parameter updates. 100,000 updates correspond to approximately half an hour, with some delays incurred due to periodic policy evaluation, which is only used for measuring the reward for the plot. One worker required significantly more than 4 hours to achieve 100% success rate, while two workers achieved the same success rate in 2.5 hours. Qualitatively, the learning process goes through a set of stages as the robots learn the task, as illustrated by learning curves in Figure 8, where the plateau near reward=0 corresponds to placing the hook near the handle, but not pulling the door open. In the first stage, the robots are unable to reach the handle, and explore the free space to determine an effective policy for reaching. Once the robots begin to contact the handle sporadically, they will occasionally pull on the handle by accident, but require additional training to be able to reach the handle consistently; this corresponds to the plateau in the learning curves. At this point, it becomes much easier for the robots to pull open the door, and a successful policy emerges. The final policy learned by the two workers was able to open the door every time, including in the presence of exploration noise.

## VII. Discussion and Future Work

We presented an asynchronous deep reinforcement learning approach that can be used to learn complex robotic manipulation skills from scratch on real physical robotic manipulators. We demonstrate that our approach can learn a complex door opening task with only a few hours of training, and our simulated results demonstrate that training times decrease with more learners. Our technical contribution consists of a novel asynchronous version of the normalized advantage functions (NAF) deep reinforcement learning algorithm, as well as a number of practical extensions to enable safe and efficient deep reinforcement learning on physical systems, and our experiments confirm the benefits of nonlinear deep neural network policies over simpler shallow representations for complex robotic manipulation tasks.

While we've shown that deep off-policy reinforcement learning algorithms are capable of learning complex manipulation skills from scratch and without purpose built representations, our method has a number of limitations. Although each of the tasks is learned from scratch, the reward function provides some amount of guidance to the learning algorithm. In the reacher task, the reward provides the distance to the target, while in the door task, it provides the distance from the gripper to the handle as well as the difference between the current and desired door pose. If the reward consists only of a binary success signal, both tasks become substantially more difficult and require considerably more exploration. However, such simple binary rewards may be substantially easier to engineer in many practical robotic learning applications. Improving exploration and learning speed in future work to enable the use of such sparse rewards would further improve the practical applicability of the class of methods explored here.

Another promising direction of future work is to investigate how diverse experience of multiple robotic platforms can be appropriately integrated into a single policy. While we take the simplest approach of pooling all collected experience, multi-robot learning differs fundamentally from single-robot learning in the diversity of experience that multiple robots can collect. For example, in a real-world instantiation of the door opening example, each robot might attempt to open a different door, eventually allowing for generalization across door types. Properly handling such diversity might benefit from explicit exploration or even separate policies trained on each robot, with subsequent pooling based on policy distillation [43]. Exploring these extensions of our method could enable the training of highly generalizable deep neural network policies in future work.

## Acknowledgements

We sincerely thank Peter Pastor, Ryan Walker, Mrinal Kalakrishnan, Ali Yahya, Vincent Vanhoucke for their assistance and advice on robot set-ups, Gabriel Dulac-Arnold and Jon Scholz for help on parallelization, and the Google Brain, X, and DeepMind teams for their support.

## References

[1] N. Kohl and P. Stone, "Policy gradient reinforcement learning for fast quadrupedal locomotion," in *International Conference on Robotics and Automation (IROS)*, 2004.
[2] G. Endo, J. Morimoto, T. Matsubara, J. Nakanishi, and G. Cheng, "Learning CPG-based biped locomotion with a policy gradient method: Application to a humanoid robot," *International Journal of Robotic Research*, vol. 27, no. 2, pp. 213–228, 2008.
[3] J. Peters and S. Schaal, "Reinforcement learning of motor skills with policy gradients," *Neural Networks*, vol. 21, no. 4, pp. 682–697, 2008.
[4] E. Theodorou, J. Buchli, and S. Schaal, "Reinforcement learning of motor skills in high dimensions," in *International Conference on Robotics and Automation (ICRA)*, 2010.
[5] J. Peters, K. Mülling, and Y. Altün, "Relative entropy policy search," in *AAAI Conference on Artificial Intelligence*, 2010.
[6] M. Kalakrishnan, L. Righetti, P. Pastor, and S. Schaal, "Learning force control policies for compliant manipulation," in *International Conference on Intelligent Robots and Systems (IROS)*, 2011.
[7] P. Abbeel, A. Coates, M. Quigley, and A. Ng, "An application of reinforcement learning to aerobatic helicopter flight," in *Advances in Neural Information Processing Systems (NIPS)*, 2006.
[8] J. Kober, J. A. Bagnell, and J. Peters, "Reinforcement learning in robotics: A survey," *International Journal of Robotic Research*, vol. 32, no. 11, pp. 1238–1274, 2013.
[9] P. Pastor, H. Hoffmann, T. Asfour, and S. Schaal, "Learning and generalization of motor skills by learning from demonstration," in *International Conference on Robotics and Automation (ICRA)*, 2009.
[10] T. P. Lillicrap, J. J. Hunt, A. Pritzel, N. Heess, T. Erez, Y. Tassa, D. Silver, and D. Wierstra, "Continuous control with deep reinforcement learning," *International Conference on Learning Representations (ICLR)*, 2016.
[11] S. Gu, T. Lillicrap, I. Sutskever, and S. Levine, "Continuous deep q-learning with model-based acceleration," in *International Conference on Machine Learning (ICML)*, 2016.
[12] M. Deisenroth, G. Neumann, and J. Peters, "A survey on policy search for robotics," *Foundations and Trends in Robotics*, vol. 2, no. 1-2, pp. 1–142, 2013.
[13] K. J. Hunt, D. Sbarbaro, R. Żbikowski, and P. J. Gawthrop, "Neural networks for control systems: A survey," *Automatica*, vol. 28, no. 6, pp. 1083–1112, Nov. 1992.
[14] M. Riedmiller, "Neural fitted q iteration–first experiences with a data efficient neural reinforcement learning method," in *European Conference on Machine Learning*. Springer, 2005, pp. 317–328.
[15] R. Hafner and M. Riedmiller, "Neural reinforcement learning controllers for a real robot application," in *International Conference on Robotics and Automation (ICRA)*, 2007.
[16] M. Riedmiller, S. Lange, and A. Voigtlaender, "Autonomous reinforcement learning on raw visual input data in a real world application," in *International Joint Conference on Neural Networks*, 2012.
[17] J. Koutník, G. Cuccu, J. Schmidhuber, and F. Gomez, "Evolving large-scale neural networks for vision-based reinforcement learning," in *Conference on Genetic and Evolutionary Computation*, ser. GECCO '13, 2013.
[18] J. Schulman, S. Levine, P. Moritz, M. Jordan, and P. Abbeel, "Trust region policy optimization," in *International Conference on Machine Learning (ICML)*, 2015.

[19] S. Levine, C. Finn, T. Darrell, and P. Abbeel, "End-to-end training of deep visuomotor policies," *Journal of Machine Learning Research (JMLR)*, vol. 17, 2016.

[20] M. Deisenroth and C. Rasmussen, "PILCO: a model-based and data-efficient approach to policy search," in *International Conference on Machine Learning (ICML)*, 2011.

[21] T. Moldovan, S. Levine, M. Jordan, and S. Abbeel, "Optimism-driven exploration for nonlinear systems," in *International Conference on Robotics and Automation (ICRA)*, 2015.

[22] R. Lioutikov, A. Paraschos, G. Neumann, and J. Peters, "Sample-based information-theoretic stochastic optimal control," in *International Conference on Robotics and Automation*, 2014.

[23] M. P. Deisenroth, G. Neumann, J. Peters *et al.*, "A survey on policy search for robotics." *Foundations and Trends in Robotics*, vol. 2, no. 1-2, pp. 1–142, 2013.

[24] Y. Chebotar, M. Kalakrishnan, A. Yahya, A. Li, S. Schaal, and S. Levine, "Path integral guided policy search," *arXiv preprint arXiv:1610.00529*, 2016.

[25] R. Williams, "Simple statistical gradient-following algorithms for connectionist reinforcement learning," *Machine Learning*, vol. 8, no. 3-4, pp. 229–256, May 1992.

[26] C. J. Watkins and P. Dayan, "Q-learning," *Machine learning*, vol. 8, no. 3-4, pp. 279–292, 1992.

[27] R. Sutton, D. McAllester, S. Singh, and Y. Mansour, "Policy gradient methods for reinforcement learning with function approximation," in *Advances in Neural Information Processing Systems (NIPS)*, 1999.

[28] J. Koutník, G. Cuccu, J. Schmidhuber, and F. Gomez, "Evolving large-scale neural networks for vision-based reinforcement learning," in *Proceedings of the 15th annual conference on Genetic and evolutionary computation*. ACM, 2013, pp. 1061–1068.

[29] V. Mnih *et al.*, "Human-level control through deep reinforcement learning," *Nature*, vol. 518, no. 7540, pp. 529–533, 2015.

[30] V. Mnih, A. P. Badia, M. Mirza, A. Graves, T. Lillicrap, T. Harley, D. Silver, and K. Kavukcuoglu, "Asynchronous methods for deep reinforcement learning," in *International Conference on Machine Learning (ICML)*, 2016, pp. 1928–1937.

[31] R. Hafner and M. Riedmiller, "Reinforcement learning in feedback control," *Machine learning*, vol. 84, no. 1-2, pp. 137–169, 2011.

[32] M. Inaba, S. Kagami, F. Kanehiro, and Y. Hoshino, "A platform for robotics research based on the remote-brained robot approach," *International Journal of Robotics Research*, vol. 19, no. 10, 2000.

[33] J. Kuffner, "Cloud-enabled humanoid robots," in *IEEE-RAS International Conference on Humanoid Robotics*, 2010.

[34] B. Kehoe, A. Matsukawa, S. Candido, J. Kuffner, and K. Goldberg, "Cloud-based robot grasping with the google object recognition engine," in *IEEE International Conference on Robotics and Automation*, 2013.

[35] B. Kehoe, S. Patil, P. Abbeel, and K. Goldberg, "A survey of research on cloud robotics and automation," *IEEE Transactions on Automation Science and Engineering*, vol. 12, no. 2, April 2015.

[36] A. Yahya, A. Li, M. Kalakrishnan, Y. Chebotar, and S. Levine, "Collective robot reinforcement learning with distributed asynchronous guided policy search," *arXiv preprint arXiv:1610.00673*, 2016.

[37] E. Todorov, T. Erez, and Y. Tassa, "Mujoco: A physics engine for model-based control," in *2012 IEEE/RSJ International Conference on Intelligent Robots and Systems*. IEEE, 2012, pp. 5026–5033.

[38] J. Ba and D. Kingma, "Adam: A method for stochastic optimization," 2015.

[39] J. Kober and J. Peters, "Learning motor primitives for robotics," in *International Conference on Robotics and Automation (ICRA)*, 2009.

[40] R. Tedrake, T. W. Zhang, and H. S. Seung, "Learning to walk in 20 minutes."

[41] S. Ioffe and C. Szegedy, "Batch normalization: Accelerating deep network training by reducing internal covariate shift," *arXiv preprint arXiv:1502.03167*, 2015.

[42] J. L. Ba, J. R. Kiros, and G. E. Hinton, "Layer normalization," *arXiv preprint arXiv:1607.06450*, 2016.

[43] A. Rusu, S. Colmenarejo, C. Gulcehre, G. Desjardins, J. Kirkpatrick, R. Pascanu, V. Mnih, K. Kavukcuoglu, and R. Hadsell, "Policy distillation," in *International Conference on Learning Representations (ICLR)*, 2016.